# A Simple CW-SSIM Kernel-based Nearest Neighbor Method for Handwritten Digit Classification

Jiheng Wang[a], Guangzhe Fan[a] and Zhou Wang[b]
[a]Dept. of Statistics and Actuarial Science, Univ. of Waterloo, Waterloo, ON, Canada
[b]Dept. of Electrical and Computer Engineering, Univ. of Waterloo, Waterloo, ON, Canada

## Abstract

We propose a simple kernel based nearest neighbor approach for handwritten digit classification. The "distance" here is actually a kernel defining the similarity between two images. We carefully study the effects of different number of neighbors and weight schemes and report the results. With only a few nearest neighbors (or most similar images) to vote, the test set error rate on MNIST database could reach about 1.5%-2.0%, which is very close to many advanced models.

## Introduction

Due to the high dimensionality nature of digital images, image classification algorithms typically require a feature extraction process (such as corner detection) or an appearance-based dimension reduction stage (such as principle component analysis) before the application of statistical learning and classification algorithms. Meanwhile, there has been some interesting recent progress on defining similarity metrics between two images that are in their original 2D functional form. These include the structural similarity (SSIM) index[1] and its extension – complex wavelet SSIM (CW-SSIM) index[2,3]. Conceptually, these similarity metrics have the potentials to be used in image classification problems, but there has not been sufficient study on how this should be performed in real-world scenarios.

Image similarity indices play a crucial role in the development, assessment and optimization of a large number of image processing and pattern recognition systems. An image can be viewed as a 2-D function of intensity. Perhaps the simplest way to compare the similarity of two images is to compute the mean squared error between these two 2D functions. Unfortunately, such a point-wise similarity measure does not take into account the correlation between neighboring image pixels and has been shown to be problematic in many ways[4]. Recently, a substantially different approach called the SSIM index[1] was proposed, where the structural information of an image is defined as those attributes that represent the structures of the objects in the visual scene, apart from the mean intensity and contrast. Thus, the SSIM index separates the comparison of local structural patterns from local mean intensity and contrast comparisons. The SSIM index has shown somewhat surprising success in predicting perceptual image quality when compared with more sophisticated methods based on psychological models of the human visual system[4]. A common drawback of both MSE and SSIM metrics is their high sensitivity to small geometric distortions such as translation, rotation and scaling. The CW-SSIM measure overcomes this problem by transforming SSIM to the complex wavelet transform domain[2,3]. The key idea behind CW-SSIM is that small geometric image distortions lead to consistent phase changes in local wavelet coefficients, and that a consistent phase shift of the coefficients does not change the structural content of the image. The potential of CW-SSIM has been demonstrated with a series of applications, including image quality assessment[2], digit recognition[2], line-drawing comparison[3], segmentation comparison[3], range-based face recognition[5] and palmprint recognition[6].

The well-known MNIST database of handwritten digits is composed of 60,000 training and 10,000 test examples, where the data were collected among Census Bureau employees and high school students. The original images have a normalized size of 28×28 and contain gray levels for the purpose of anti-aliasing. In this paper we propose a series of simple and fast kernel-based classification algorithms based on CW-SSIM index for the MNIST database, which appears to be effective and reliable tools for the MNIST Database of Handwritten Digits Classification. Although no feature extraction or dimension reduction process is involved, we obtain quite competitive results with only a simple *k*-NN model. Given that the CW-SSIM index provides a powerful similarity measure between two misaligned images and there are sufficient training examples in the MNIST database, we are able to effectively classify test samples using only the most similar images.

# Methodology

## Complex Wavelet Structural Similarity Index (CW-SSIM)

The SSIM index was originally proposed to predict perceived image quality[1,4]. The fundamental principle is that the human visual system is highly adapted to extract structural information from the visual scene, and therefore, a measurement of structural similarity should provide a good approximation of perceptual image quality. In particular, SSIM attempts to discount those distortions that do not affect the structures (or local intensity patterns) of the image. In the spatial domain, the SSIM index between two image patches $x = \{x_i \mid i = 1, 2, \ldots, M\}$ and $y = \{y_i \mid i = 1, \ldots, M\}$ is defined as

$$S(x,y) = \frac{(2\mu_x\mu_y + C_1)(2\sigma_{xy} + C_2)}{(\mu_x^2 + \mu_y^2 + C_1)(\sigma_x^2 + \sigma_y^2 + C_2)} \tag{1}$$

where $\mu$ and $\sigma$ are the sample mean, standard deviation and covariance terms of $x$, $y$ and $xy$, respectively, and $C_1$ and $C_2$ are two small positive constants to avoid instabilities. The maximum value 1 is achieved if and only if $x$ and $y$ are identical.

The major drawback of the spatial domain SSIM algorithm is that it is highly sensitive to translation, scaling, and rotation of images. The CW-SSIM index is an extension of the SSIM method to the complex wavelet domain. The goal is to design a measurement that is insensitive to “non-structural” geometric distortions that are typically caused by nuisance factors, such as changes in lighting conditions and the relative movement of the image acquisition device, rather than the actual changes in the structures of the objects. The CW-SSIM index is also inspired by the impressive patter recognition capabilities of the human visual system[1]. In the last three decades, scientists have found that neurons in the primary visual cortex can be well-modeled using localized multi-scale bandpass oriented filters that decompose natural image signals into multiple visual channels. Interestingly, some psychophysical evidence suggests that the same set of visual channels may also be used in image pattern recognition tasks[7]. Furthermore, phase contains more structural information than magnitude in typical natural images, and rigid translation of image structures leads to consistent phase shift. The CW-SSIM index is defined as

$$\tilde{S}\left(c_x, c_y\right) = \frac{2\left|\sum_{i=1}^{N} c_{x,i} c_{y,i}^{*}\right| + K}{\sum_{i=1}^{N}\left|c_{x,i}\right|^2 + \sum_{i=1}^{N}\left|c_{y,i}\right|^2 + K} \tag{2}$$

Here $c_x$ and $c_y$ are the sets of local coefficients (in the neighboring spatial locations of the same wavelet subband) extracted from the complex wavelet transformation (e.g. the complex version of the steerable pyramid decomposition[8]) of the two images being compared, respectively, $c^*$ denotes the complex conjugate of $c$, $K$ is a small positive constant. The purpose of $K$ is mainly to improve the robustness of the CW-SSIM measure when the local signal-to-noise ratios are low. We consider CW-SSIM as a useful measure of image structural similarity based on the beliefs that 1) the structural information of local image features is mainly contained in the relative phase patterns of wavelet coefficients, and 2) constant phase shift of all coefficients does not change the structure of the local image feature.

## K-Nearest Neighbors Algorithm (K-NN)

In pattern recognition, the k-nearest neighbors algorithm (K-NN) is a method for classifying objects based on closest training examples in the feature space. K-NN is a type of instance-based learning, or lazy learning where the function is only approximated locally and all computation is deferred until classification. The k-nearest neighbor algorithm is amongst the simplest of all machine learning algorithms: an object is classified by a majority vote of its neighbors, with the object being assigned to the class most common amongst its $k$ nearest neighbors ($k$ is a positive integer, typically small). If $k = 1$, then the object is simply assigned to the class of its nearest neighbor.

## The Classification Algorithm for MNIST Database

Here we will introduce a series of simple kernel-based classification algorithms for the MNIST database, which uses CW-SSIM index to define the distance between different images. A detailed and systematic description of kernel methods for machine learning is given in Schölkopf and Smola[9]. As we stated before, only employing a basic statistical model, i.e. $k$-NN, we will only use most similar images to classify test examples.

We select 10 test images randomly, one for each digit, and show their CW-SSIM scores with all training images in Figure 1. It can be observed that the value of CW-SSIM is typically higher when it is used to measure the similarity between two images belonging to the same digit, which verifies our hypothesis of using most similar images to classify.

### K=1 Case

When *k*=1, this is the most direct implementation: we classify every test image using the most similar training image, i.e., we choose the training image with the highest CW-SSIM score to assign the label. Given that there are 60,000 examples in the training set, this method is reasonable since the training set is large enough to provide the most similar image to every test example.

### K>1 Case

When *k*>1, an image is classified by a majority vote of its neighbors. For the unweighted case, we denote by the most frequent digit among the most *k* similar images as our predicted label. While for the weighted case, we use the corresponding CW-SSIM scores as the weights of these *k* similar images. Then the digit with the highest sum of weights (CW-SSIM scores) is denoted as the predicted label.

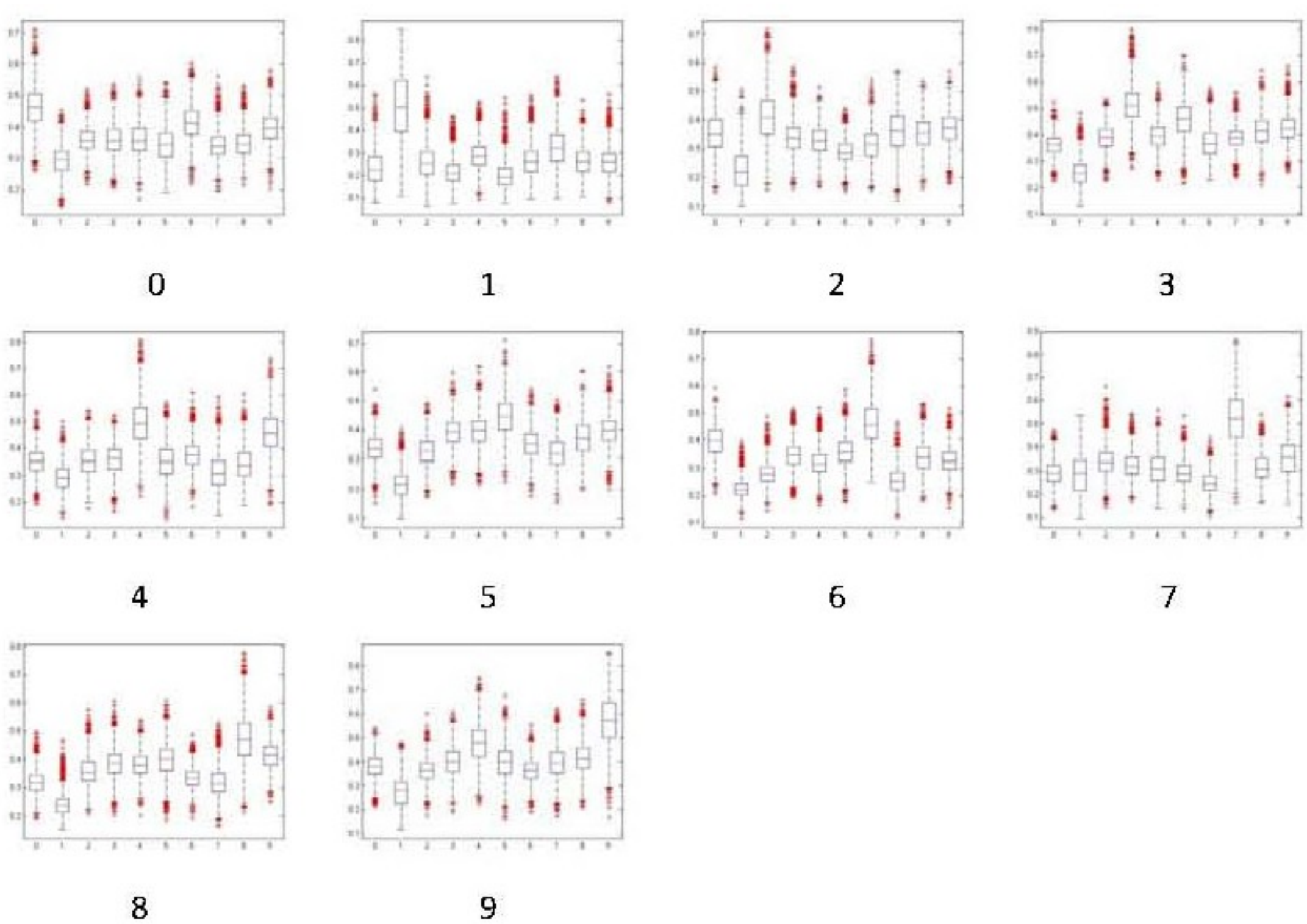

**Figure 1. The boxplots of CW-SSIM scores (One test sample is selected for each digit)**

# Result

## Experiment on Simulated data

The experiment is first carried out on simulated data. A training/testing digit image database of 2,000 images was created by shifting, scaling, rotating, and blurring ten hand-written template digit images. Figure 2 shows a random subset of examples in our image database.

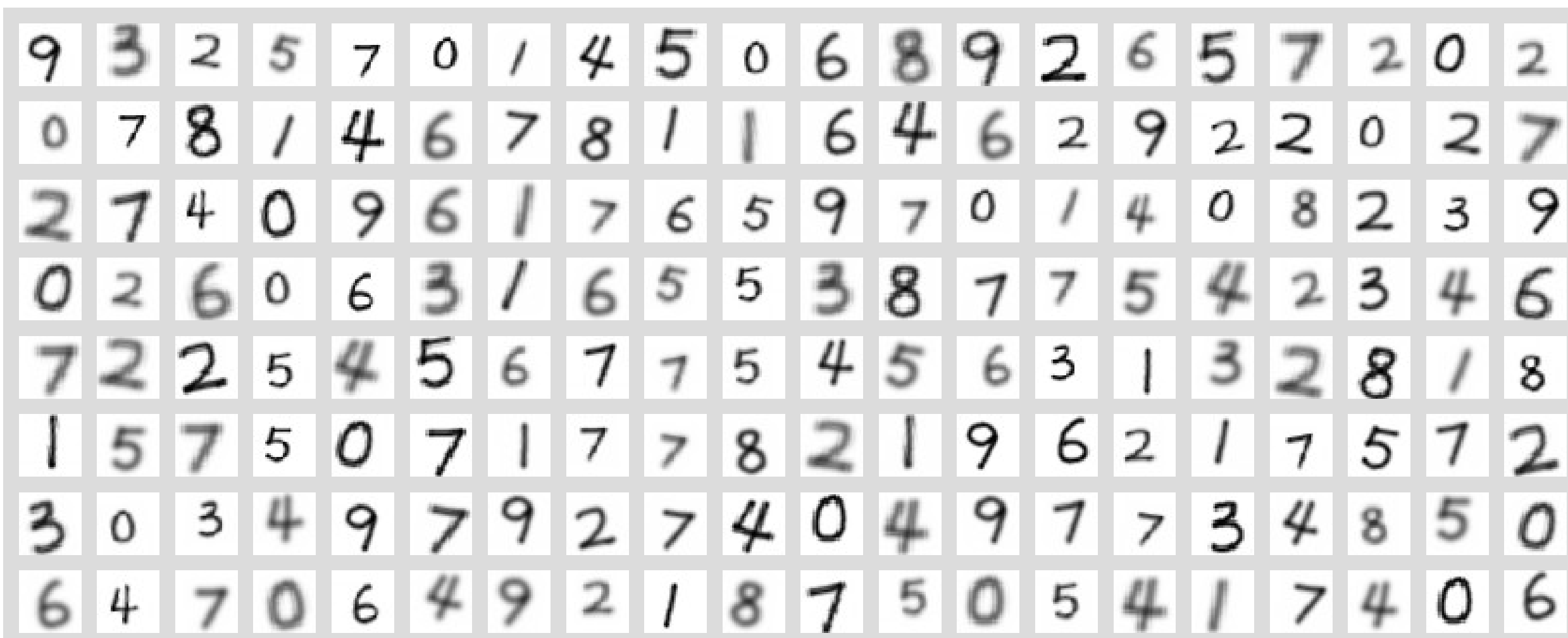

**Figure 2. Random samples of simulated hand-written digit images**

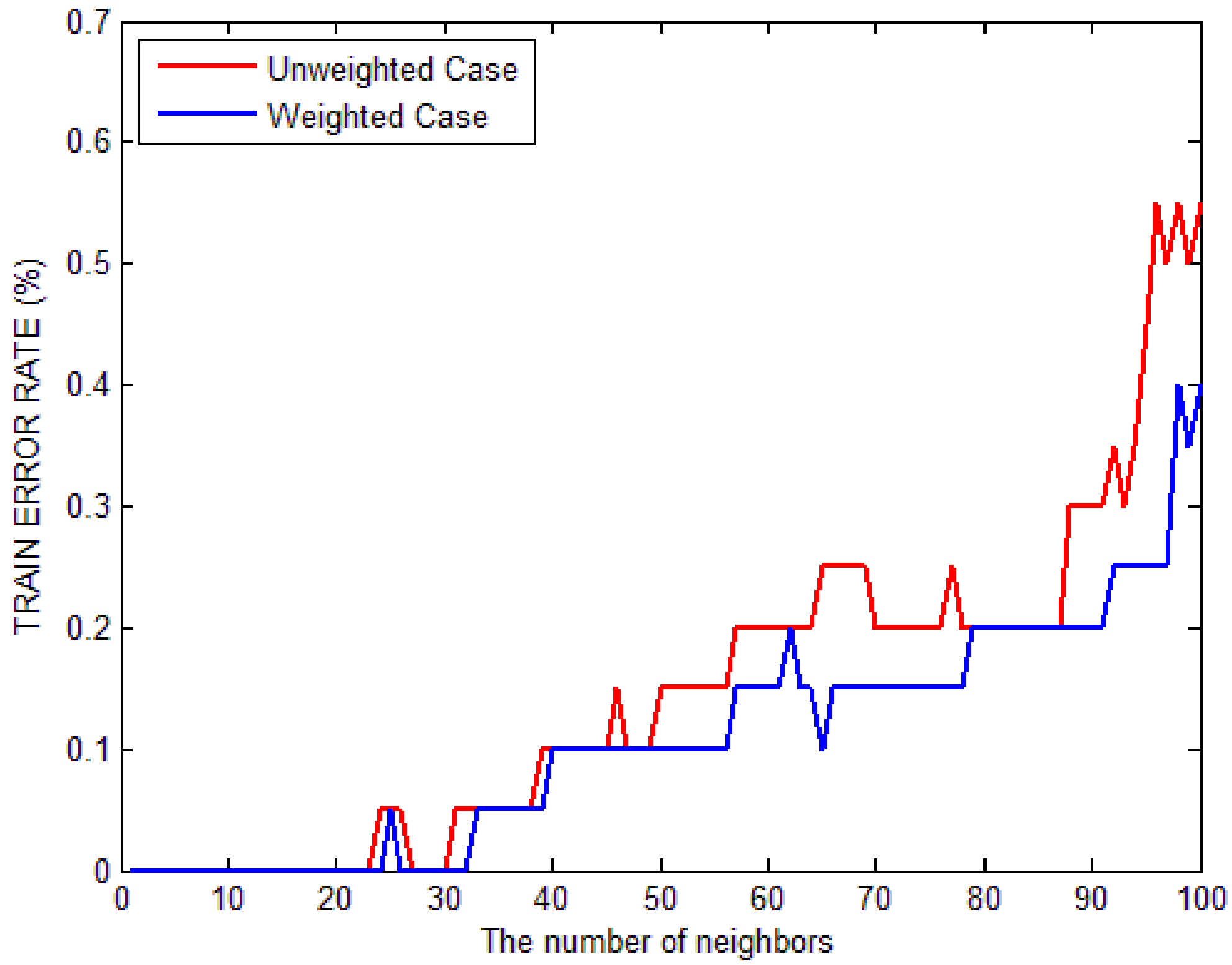

**Figure 3. Training error of CW-SSIM on Simulated data using 2,000 training images as a function of the number of neighbors *k***

We repeated 5 trials of tests where 1,600 images randomly selected from 2,000 images were used for training and the remaining 400 images for testing.The test results are shown in Figure 3, including the unweighted and weighted cases. The training error rate, as a function of the number of neighbors, is computed as the average percentage of misclassified images for each trial over 2,000. When $k \leq 20$, the training error rates for the unweighted or the weighted case are all zero. When $k>20$, the train error rates vary from 0 to 0.6%. This result shows that we can obtain a higher predicted accuracy when $k$ is not too large, which is in accordance with our proposal of only using most similar images to classify.

## Experiment on MNIST database

The MNIST database is composed of 60,000 training and 10,000 test examples. Some sample images are shown in Figure 4. In our preliminary test, we extracted a subset from the database for training and testing.

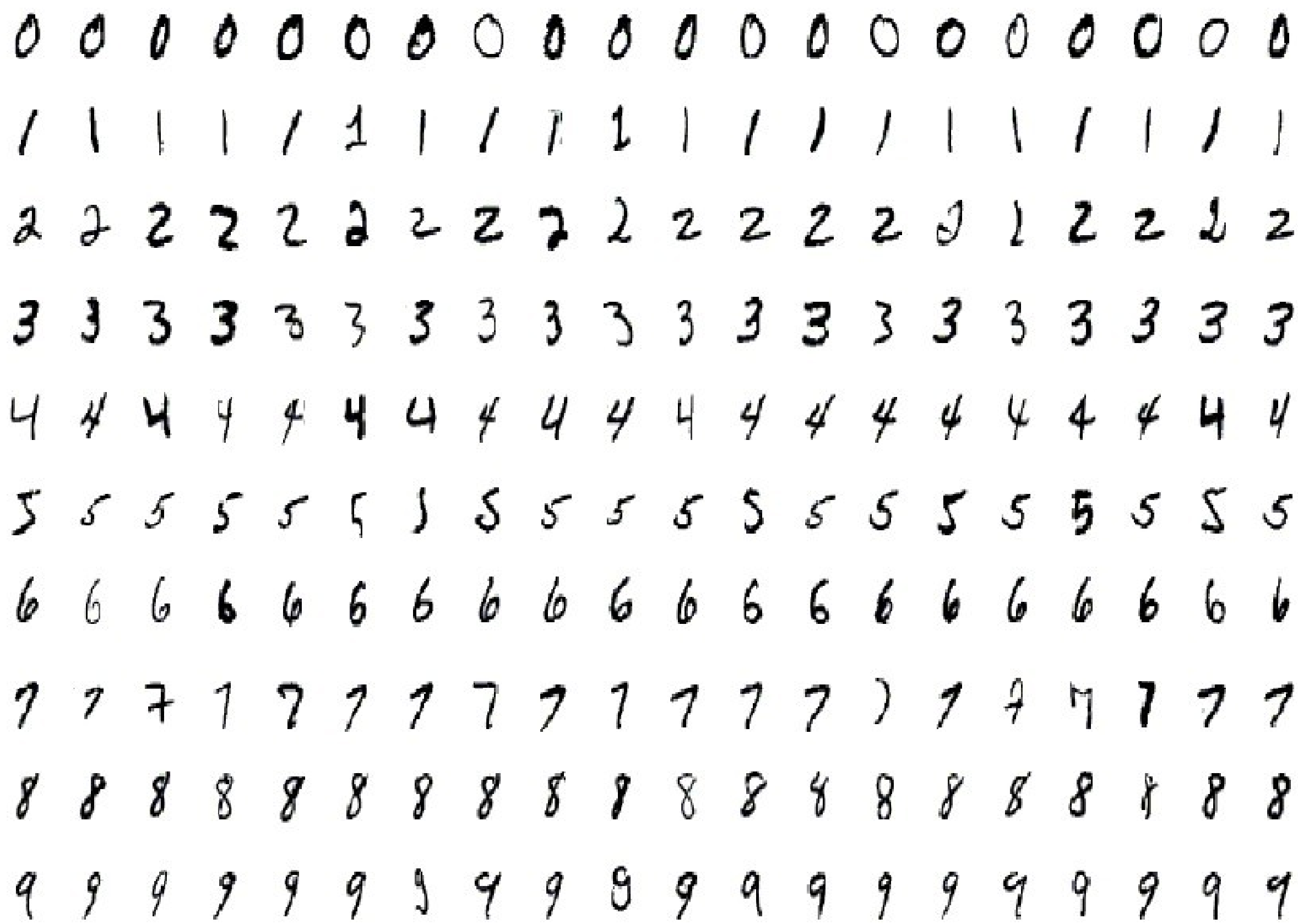

**Figure 4. Random samples of hand-written digit images from the MNIST database**

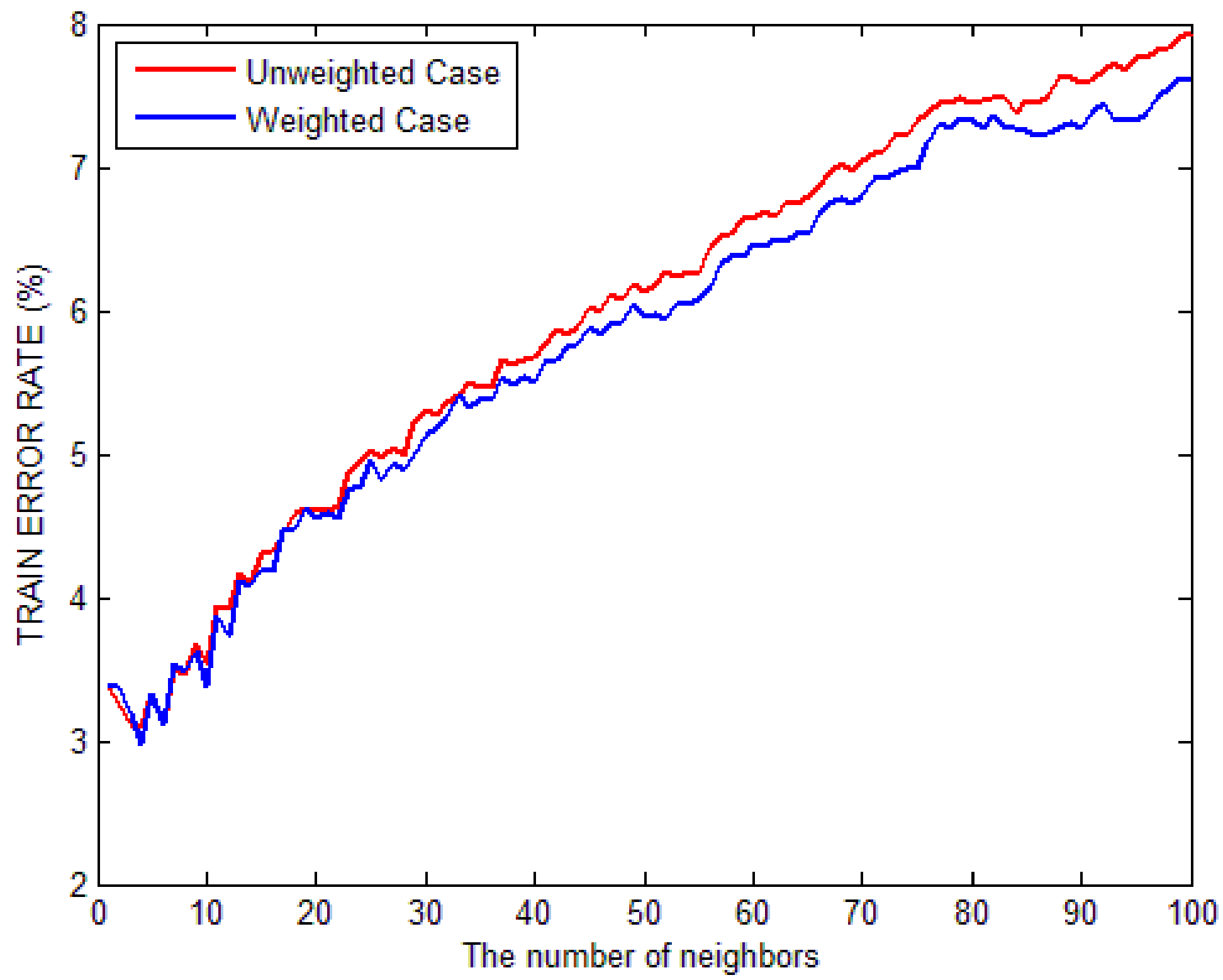

**Figure 5. Training error rate as a function of the number of neighbors *k* for CW-SSIM on MNIST database using 5,000 training images**

First, we selected 5,000 training images and 2,000 test images randomly. We repeated 5 trials of training tests where 4,000 images randomly selected from 5,000 images were used for training and the remaining 1,000 images for testing. The test results are shown in Figure 5, including the unweighted and weighted cases. The training error rate is computed as the average percentage of misclassified images for each trial over 5,000. When $k$=1, the train error rate is 3.38%. When $k$>1, for the unweighted case, the lowest error rate is 3.10% when using 4 neighbors and the second lowest error rate is 3.12% when using 3 or 6 neighbors. While for the weighted case, the lowest error rate reaches 2.98% when using 4 neighbors and the error rates are above 3.00% when using other choices. For both cases, the training error rate is always below 4.00% when using 2 to 12 neighbors. Roughly speaking, the weighted method performs slightly better than the unweighted one.

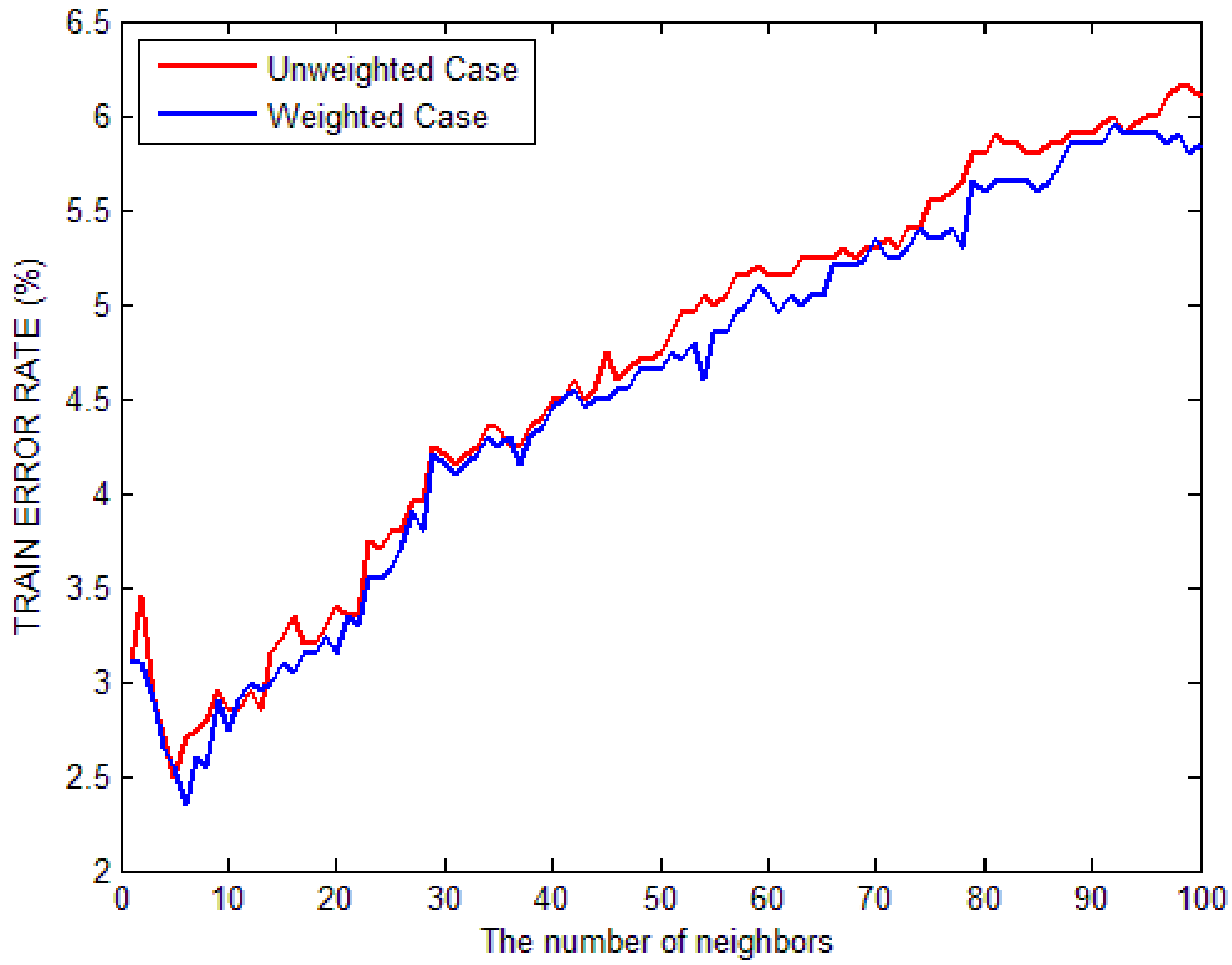

**Figure 6. Test error rate as a function the number of neighbors *k* for CW-SSIM on MNIST database using 5,000 training images and 2,000 test images**

The results in Figure 6 used all the 5,000 images for training and the separate set of 2,000 images for testing. The test error rate is calculated as the average percentage of misclassified images over 2,000. When $k$=1, the test error rate is 3.10%. When $k$>1, for the unweighted case, the lowest error rate is 2.50% when using 5 neighbors and the error rate is always below 3.00% when using 3 to 13 neighbors. While for the weighted case, the lowest error rate reaches 2.35% when using 6 neighbors and the error rate is also below 3.00% when using 3 to 11 neighbors. Similar to the training process described above, the weighted method performs slightly better than the unweighted one.

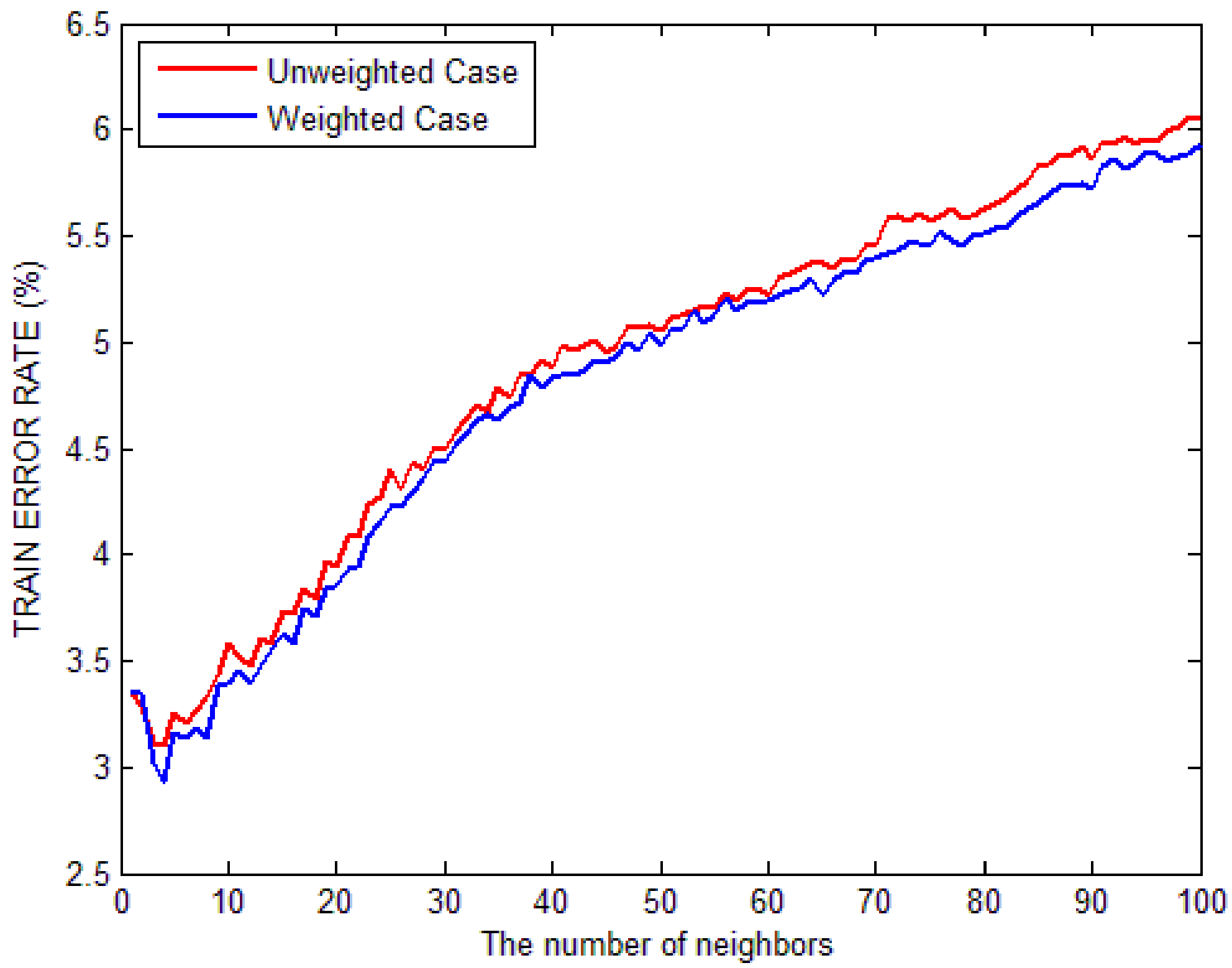

**Figure 7. Training error rate as a function of the number of neighbors *k* for CW-SSIM on MNIST database using 10,000 training images**

In our second test, we selected 10,000 training images and 5,000 test images randomly. Similarly, we repeated 5 trials of tests where 8,000 images randomly selected from 10,000 images were used for training and the remaining 2,000 images from the same set were employed for testing. The test results are shown in Figure 7, including both cases. The results in Figure 7 used all 10,000 images for training and a separate set of 2,000 images for testing. When $k$=1, the test error rate is 3.35%. When $k$>1, for the unweighted case, the lowest error rate is 3.10% when using 4 neighbors and the second lowest rate is 3.11% when using 3 neighbors. While for the weighted case, the lowest error rate reaches 2.93% when using 4 neighbors and the second lowest rate is 3.02% when using 3 neighbors. The weighted method is slightly better than the unweighted one.

The results in Figure 8 used all the 10,000 images for training and a separate set of 5,000 images for testing. The test error rate is calculated as the average percentage of misclassified images over 5,000. When $k$=1, the test error rate is 4.94%. When $k$>1, for the unweighted case, the lowest error rate is 4.30% when using 4 neighbors and the second lowest error rate is 4.48% when using 6 neighbors. While for the weighted case, the lowest error rate is 4.42% when using 5 or 6 neighbors and the second lowest error rate is 4.50% when using 8 neighbors. There is no significant difference between weighted and unweighted methods.

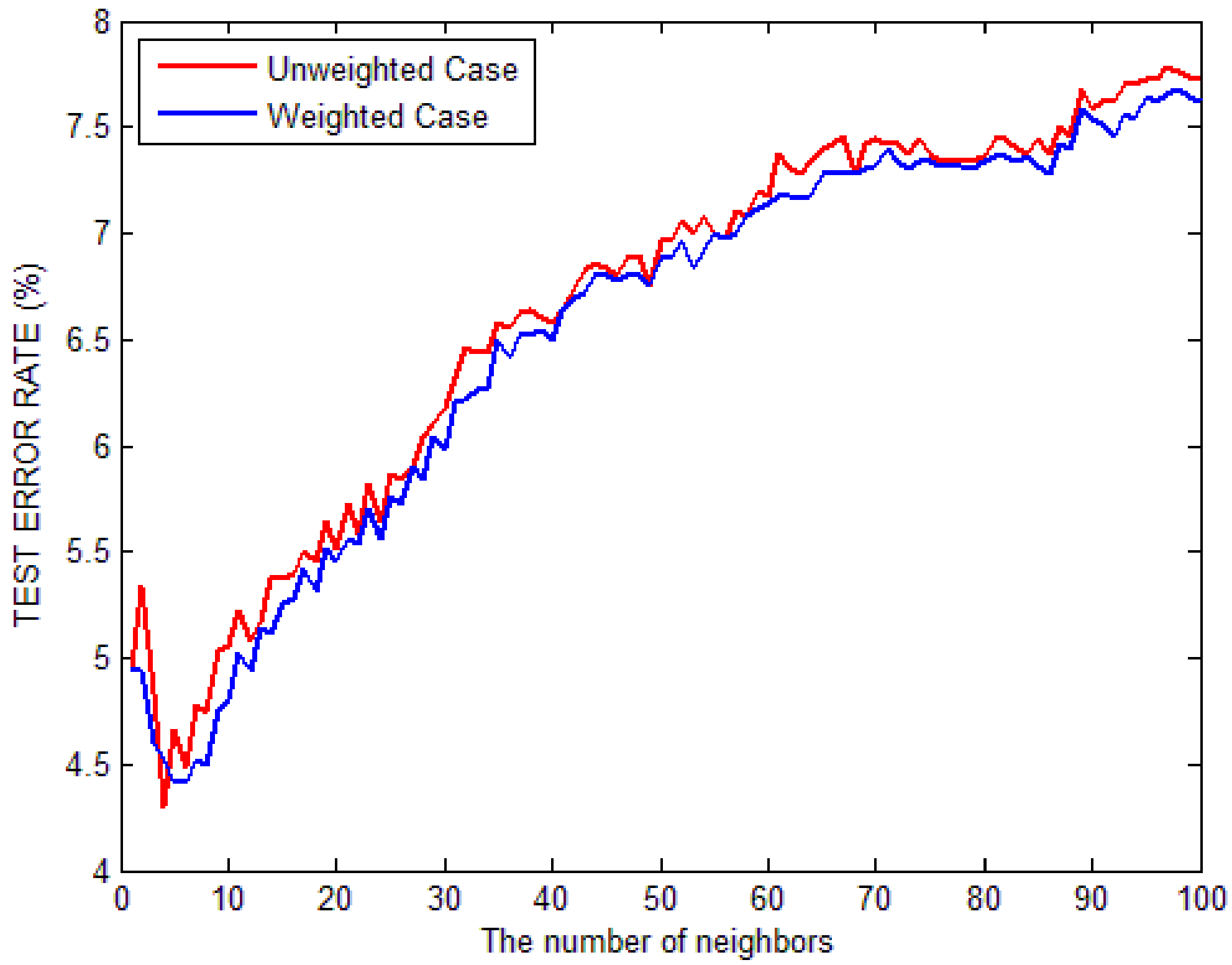

**Figure 8. Test error rage as a function of the number of neighbors *k* for CW-SSIM on MNIST database using 10,000 training and 5,000 test images**

From the above two training tests, there are two observations. First, there is no significant difference between weighted and unweighted methods although some results showed that the weighted one performs slightly better. Second, it is hard to decide the exact number of neighbors. A choice between 4 to 10 neighbors seems to be acceptable, which is in accordance with our proposal of only using most similar images to classify.

Finally, we computed the case for 60,000 training images and 10,000 test images. This test is costly in terms of computational. However, compared with many existing machine learning techniques, the speed of our algorithms are reasonable. The results are shown in Figure 9.

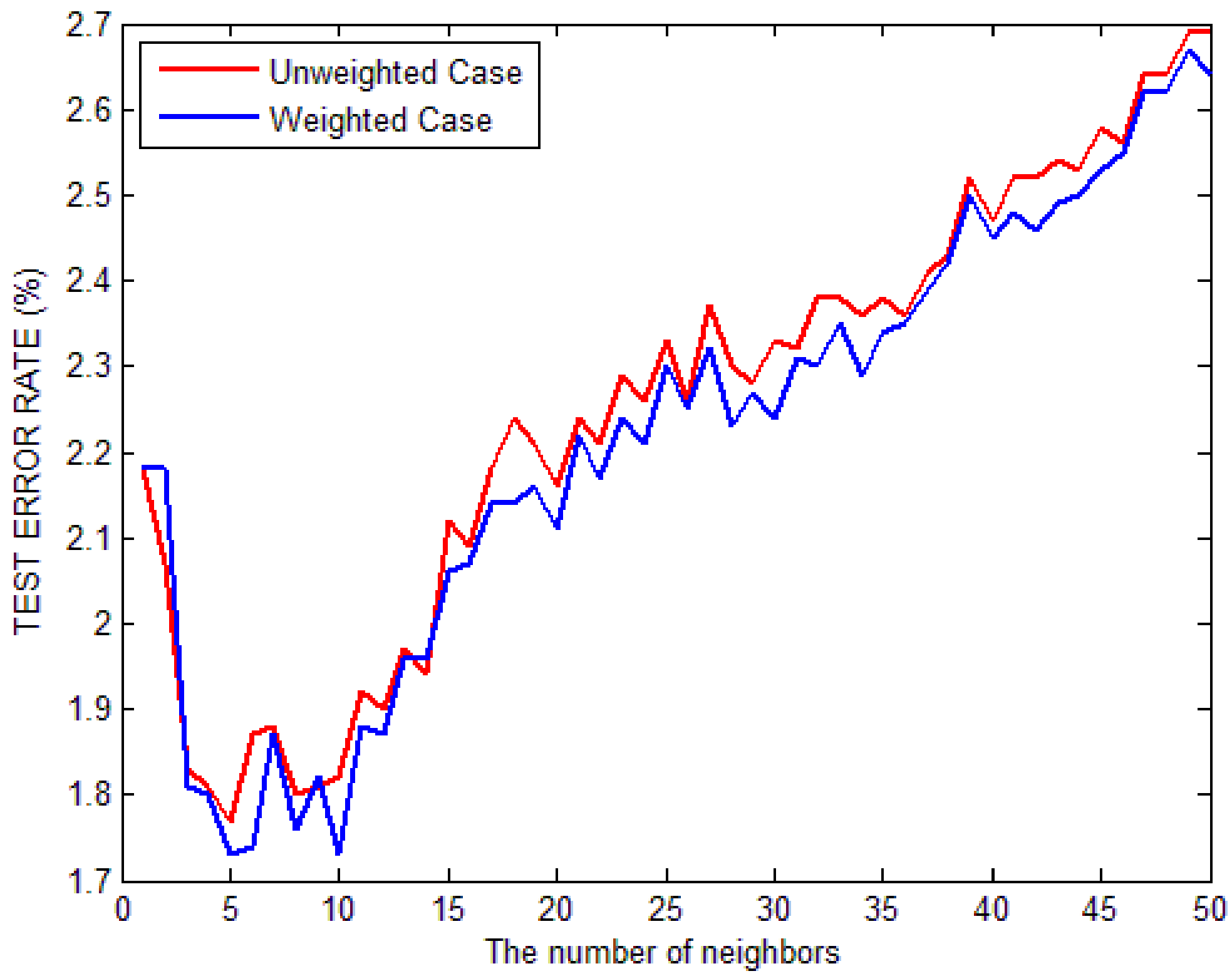

**Figure 9. Test error rate as a function of neighbors *k* for CW-SSIM on MNIST database using 60,000 training and 10,000 test images**

When *k*=1, the test error rate is 2.18%. When *k*>1, for the unweighted case, the test error rate is below 2% when using 3 to 14 neighbors and the lowest error rate is 1.77% when using 5 neighbors. While for the weighted case, the result is quite similar to the unweighted one, with the lowest error rate reaching 1.73% when using 5 or 10 neighbors. There is no significant difference between weighted and unweighted methods.

Some test images and their most similar training images are shown in Figure 10. It can be observed that there exist some "bad" training images that could be quite misleading and severely reduce the prediction accuracy. Thus how to eliminate these bad images from the training set could be a promising direction for future research.

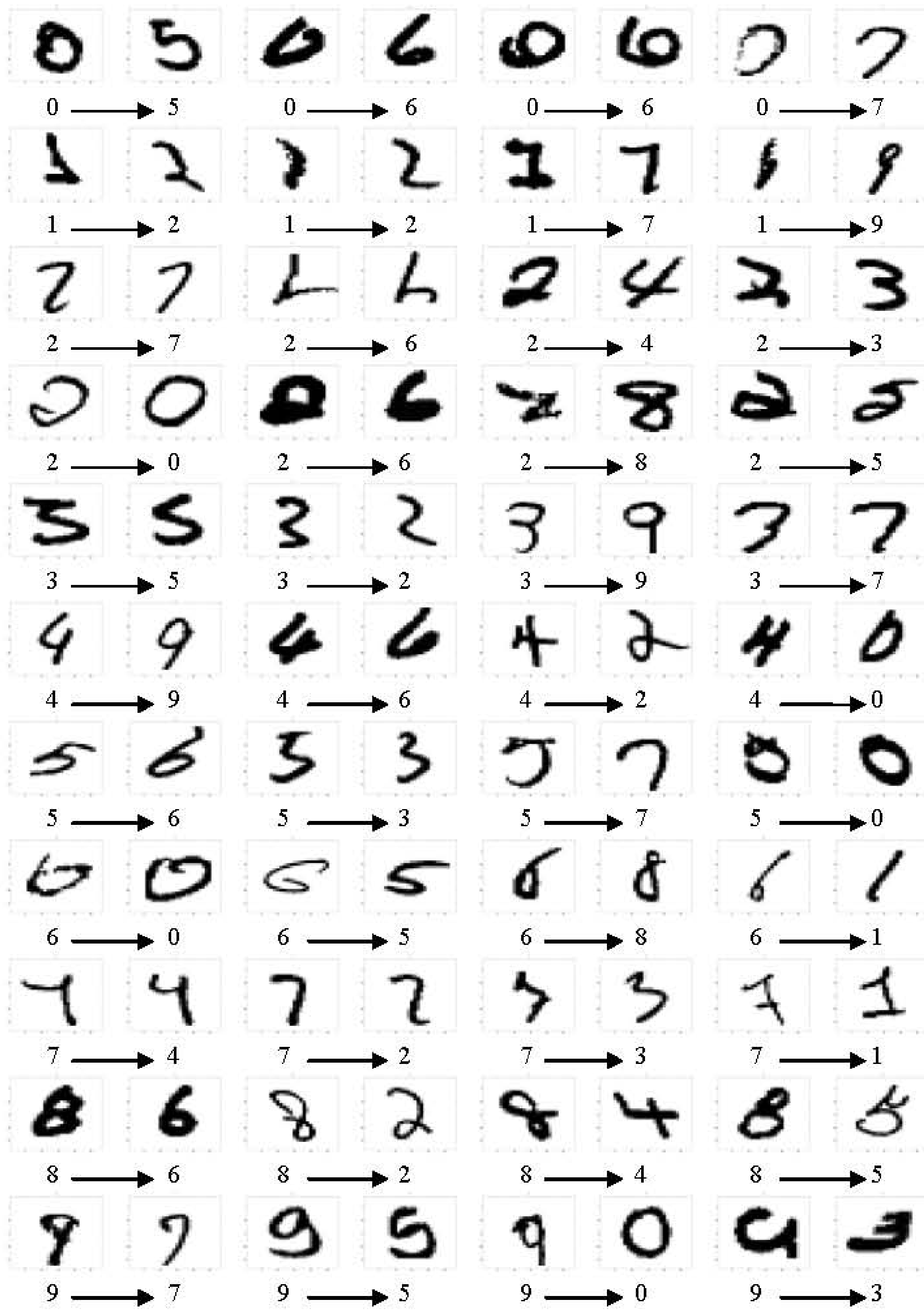

**Figure 10. Some test examples and their most similar images in the training set**

## Further discussion

### Introducing Decaying Weighting Function

Since it is hard to decide the exact number of neighbors, we introduce a smooth decaying weighting function. In particular, we attempted an exponential function and a Gaussian function given by

$$w(i) = e^{-\frac{i-1}{\sigma}}$$

$$w(i) = e^{-\frac{(i-1)^2}{\sigma}} \quad (3)$$

where $i$ is the rank according to the CW-SSIM scores in descending order. Based on the weighted method, we use the product of this additional decaying weight *w(i)* and the CW-SSIM scores as the weights of all training images, and the digit with the highest sum of weights is denoted as the predicted label. Both decaying functions in Eq. (3) are tested and the results are shown in Figure 11.

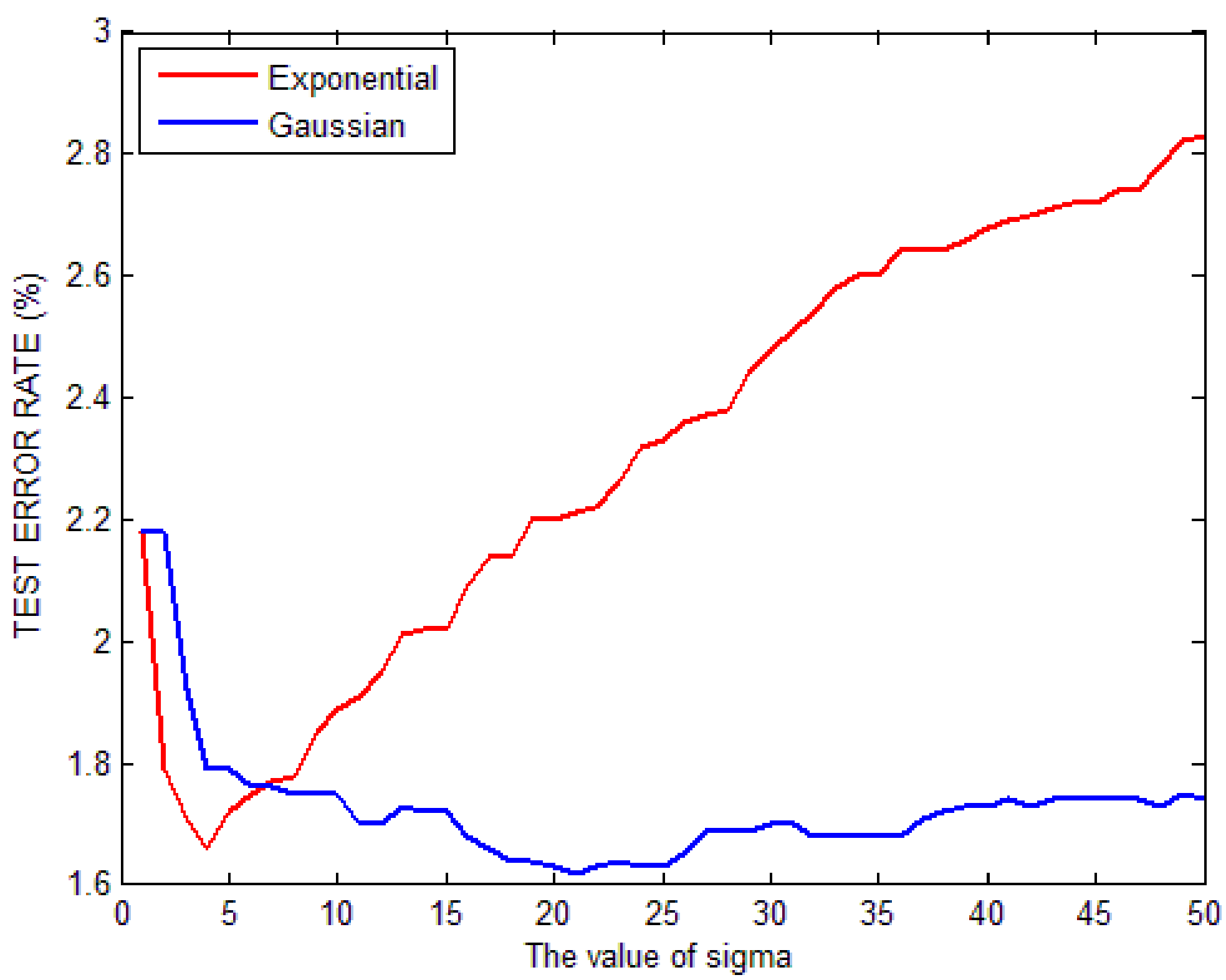

**Figure 11. Test error of CW-SSIM on MNIST database as a function of the value of $\sigma$**

When using an exponential decaying function, the lowest error rate is 1.66% with $\sigma$ set to 4. While using a Gaussian decaying function, the lowest error rate reaches 1.62% with $\sigma$ set to 21. From the two curves in Figure 11, it seems Gaussian decaying function is a more reliable and stable choice especially, which gives consistent results within a range of $\sigma$ values (from 20 to 25). The error rate of about 1.62% within that range is the best result of all methods we tested so far.

## Comparison

The results of all three algorithms are shown in Table 1, where the method with decaying weighting function involved results in to the best prediction accuracy.

**Table 1. Comparison between MSI, V-MSI and Exp-MSI**

| Method | Best for $k = 1$ | Best for $k > 1$ | Best with decaying weighting function |
|---|---|---|---|
| Error rate | 2.18% | 1.73% | 1.62% |

## Conclusion

We have proposed a series of simple CW-SSIM kernel-based classification algorithms, which appears to be effective and reliable tools when tested with the MNIST Database of handwritten digit classification. An interesting feature of our approach is that no feature extraction or dimension reduction process is involved. We obtain competitive results with only a pretty simple model. This result could be understood from two aspects. First, the CW-SSIM index provides a powerful similarity measure between two misaligned images. Second, the training set is large enough to almost always provide good matching images to any test sample.